\documentclass[letterpaper, 10 pt, conference]{ieeeconf}

\IEEEoverridecommandlockouts

\usepackage{etextools}
\usepackage{amssymb}
\usepackage{float}
\usepackage{makeidx}
\usepackage{amsmath}
\usepackage{amsfonts}
\usepackage{bbm}
\usepackage{epsfig}
\usepackage{epsf}
\usepackage{psfrag}
\usepackage{verbatim}
\usepackage{multirow}
\usepackage{tabularx}
\usepackage{booktabs}
\usepackage[tight,footnotesize]{subfigure}
\usepackage{array}
\usepackage{soul}
\usepackage{footnote}
\usepackage{cite}
\usepackage{dblfloatfix}

\usepackage{color, colortbl}
\usepackage[colorlinks,bookmarksnumbered,citecolor=orange,urlcolor=orange]{hyperref}
\usepackage{graphicx}
\graphicspath{{Figures}}
\DeclareGraphicsExtensions{.pdf,.png}
\usepackage{bigstrut}
\usepackage[english]{babel}

\usepackage{csquotes}

\usepackage[T1]{fontenc}
\usepackage{algorithm, setspace}
\usepackage{algpseudocode}
\usepackage{url}
\usepackage{multirow}
\usepackage{float}
\usepackage{xcolor}
\usepackage{hyperref}
 \hypersetup{
     colorlinks=true,
     linkcolor=orange,
     filecolor=orange,
     citecolor=orange,      
     urlcolor=orange,
     }
\let\labelindent\relax
\usepackage{enumitem}

\newcommand{\p}[1]{\smallskip \noindent \textbf{{#1}.}}
\newcommand{\eq}[1]{Equation~(\ref{eq:#1})}
\newcommand{\fig}[1]{Figure~\ref{fig:#1}}
\usepackage{balance}

\title{\LARGE \bf
From Local Corrections to Generalized Skills: 
\\
Improving Neuro-Symbolic Policies with MEMO
}

\author{Benjamin A. Christie$^1$, Yinlong Dai$^1$, Mohammad Bararjanianbahnamiri$^1$,\\Simon Stepputtis$^2$, and Dylan P. Losey$^1$\vspace{-0.5em}
\thanks{This work is supported in part by NSF Grant $\#2337884$. \newline $^1$\href{https://collab.me.vt.edu/}{Collab}, Dept. of Mechanical Engineering, Virginia Tech, Blacksburg, USA. \newline $^2$\href{https://tealab.ai/}{TEA Lab}, Dept. of Mechanical Engineering, Virginia Tech, Blacksburg, USA. Email: \texttt{\{benc00, daiyinlong\}@vt.edu}}
}

\begin{document}

\maketitle
\thispagestyle{empty}
\pagestyle{empty}

%%%%%%%%%%%%%%%%%%%%%%%%%%%%%%%%%%%%%%%%%%%%%%%%%%%%%%%%%%%%%%%%%%%%%%%%%%%%%%%%
\begin{abstract}

Recent works use a neuro-symbolic framework for general manipulation policies.
The advantage of this framework is that --- by applying off-the-shelf vision and language models --- the robot can break complex tasks down into semantic subtasks.
However, the fundamental bottleneck is that the robot needs skills to ground these subtasks into embodied motions.
Skills can take many forms (e.g., trajectory snippets, motion primitives, coded functions), but regardless of their form \textit{skills act as a constraint}.
The high-level policy can only ground its language reasoning through the available skills; if the robot cannot generate the right skill for the current task, its policy will fail.
We propose to address this limitation --- and dynamically expand the robot's skills --- by leveraging user feedback.
When a robot fails, humans can intuitively explain what went wrong (e.g., ``no, go higher'').
While a simple approach is to recall this exact text the next time the robot faces a similar situation, we hypothesize that by collecting, clustering, and re-phrasing natural language corrections across multiple users and tasks, we can synthesize more general text guidance and coded skill templates.
Applying this hypothesis we develop Memory Enhanced Manipulation (MEMO).
MEMO builds and maintains a retrieval-augmented \textit{skillbook} gathered from human feedback and task successes.
At run time, MEMO retrieves relevant text and code from this skillbook, enabling the robot's policy to generate new skills while reasoning over multi-task human feedback.
Our experiments demonstrate that using MEMO to aggregate local feedback into general skill templates enables generalization to novel tasks where existing baselines fall short. 
See supplemental material here: \url{https://collab.me.vt.edu/memo}

\end{abstract}

%%%%%%%%%%%%%%%%%%%%%%%%%%%%%%%%%%%%%%%%%%%%%%%%%%%%%%%%%%%%%%%%%%%%%%%%%%%%%%%%

\section{Introduction} \label{sec:intro}

\begin{figure}[t]
    \centering
    \includegraphics[width=1.0\linewidth]{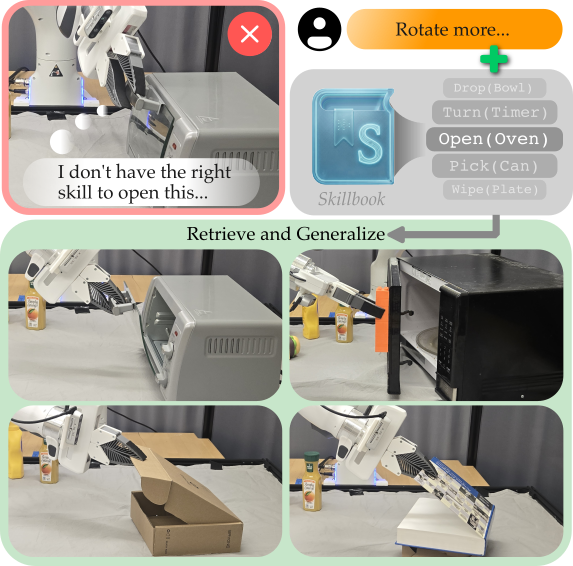}
    \vspace{-1.5em}
    \caption{Neuro-symbolic policy tries to ``toast food.'' (Top Left) The robot fails because it lacks the necessary skill for opening the toaster. (Top Right) A human gives feedback. MEMO collects this and other feedback into a retrieval-augmented skillbook, and then clusters the skillbook to extract generalized text and code templates. (Bottom) The generalized entries guide the policy's code generation for new skills, e.g., an \texttt{open\_door} skill.}
    \label{fig:front}
    \vspace{-1.5em}
\end{figure}

Neuro-symbolic approaches are a promising way to obtain robot policies by combining the benefits of neural reasoning and control.
Imagine that a human instructs the robot in \fig{front} to ``toast food.''
Under neuro-symbolic paradigms, the robot first applies a general-purpose foundation model to determine the necessary steps (e.g., ``open the door'').
The robot then tries to ground this natural language instruction into numerical control parameters (e.g., a sequence of positions and velocities).
Vision and language models excel at the first part --- identifying the right high-level subtask --- but language is ambiguous, and it is difficult to convert language into precise robot motions.

To address this problem, existing works equip the robot with \textit{skills}.
Skills are motion primitives or simple policies that are associated with semantic characteristics and interpretable parameters \cite{intelligence2025pi_, dai2026languagemovementprimitivesgrounding, kwon2024language}.
For instance, a ``grasp'' skill could control the robot arm to move to a target location and then close its gripper.
Skills are an effective way to ground the robot's policy because (a) the semantics associated with skills make it easy for the vision-language model to determine which skills to apply, and (b) the model can intuitively specify the small number of interpretable parameters that each skill requires.
Although recent research has developed multiple ways in which robots can collect or generate skill libraries \cite{liang2022code, wang2023voyager, 10161317, saycan2022arxiv}, these \textit{skills are still inherent constraints}.
If the robot does not know (or cannot generate) the right skills for a given task, the policy is unable to complete that task.

To address the central challenge of skill libraries we introduce \textbf{MEMO}, a method that enables robots to dynamically expand the ways they ground high-level semantics into low-level actions by building new generalized skills from feedback.
Consider \fig{front}.
When the robot makes a mistake, the human gives a natural language correction (e.g., ``no, you need to rotate more'').
Reasoning over this specific feedback helps the robot adapt its existing motions to their current context \cite{zha2024distilling, huang2022inner, lynch2023interactive}.
But for long-term improvements, our hypothesis is that:
\begin{center}
    \textit{By aggregating natural language corrections from multiple users across multiple tasks into a \emph{skillbook}, we can create and refine a continuously evolving set of generalized skills, leading to improvements beyond the corrected tasks}.
\end{center}
A \textit{skillbook} is a specific instance of retrieval-augmented generation (RAG) that we design for neuro-symbolic robot policies.
The skillbook begins as a set of text entries, where each entry includes the intended subtask, state information, and the human's correction for that subtask.
At run time the vision and language model retrieves relevant entries from the skillbook and then uses those entries to augment its reasoning for the current conversation.
Returning to our example: when pulling open the toaster door, the robot remembers that it should ``rotate more.''

But the fundamental advantage of MEMO is not targeted recall; it is how we evolve the skillbook over time.
As the skillbook accumulates user feedback and task successes, MEMO utilizes a language model in an asynchronous background process to cluster related text entries.
By reasoning over these clusters, MEMO refines the skillbook into parameterized code templates and generalized human corrections, which serve as a reference when the policy generates code for new skills.
See \fig{front}: by clustering several entries on how to open doors, the robot generates and refines a more invariant \texttt{open\_door()} function parameterized by the handle pose and door dimensions.
Now --- instead of retrieving specific natural language corrections --- the robot retrieves relevant code templates that go beyond the limits of both its existing skills and the code the model could have generated without any human guidance.
By aggregating feedback into generalized functions, MEMO thus provides a meaningfully expanding set of skills that neuro-symbolic policies can apply to convert language into actions.

Overall, this work is a step towards general-purpose robots that leverage human feedback to improve long-term capabilities.
We make the following contributions:

\p{Collecting and Retrieving Feedback}
We introduce a skillbook, a database containing human feedback and robot code.
MEMO automatically paraphrases human corrections into task-specific and task-invariant entries, and stores these entries alongside any code templates the robot used to successfully complete its subtasks.
Skillbook entries can be retrieved when the robot faces similar contexts.

\p{Clustering Feedback around Skill Templates}
We cluster skillbook entries while conditioning on code templates.
This process removes repetitive or contradictory feedback while also summarizing multiple human corrections across different contexts to reach more general guidance.
At run time, the robot's policy retrieves and reasons over this generalized feedback in order to generate code for new skills.

\p{Improving Beyond Local Feedback}
We collect a skillbook from user feedback across simulated tasks, and then compare MEMO to state-of-the-art baselines in simulation and the real world.
Across previously unseen tasks MEMO achieves a higher zero-shot success rate than either robotics foundation models or neuro-symbolic approaches that only reason about relevant human feedback.

\section{Related Work} \label{sec:related}

MEMO combines the common-sense reasoning capabilities of large foundation models with a continuously expanding library of skills.
We create these skills by aggregating human feedback into a skillbook and then leveraging a retrieval-augmented generation approach.

\subsection{Foundation Models for Robot Learning}

Robotics vision-language-action (VLA) models such as RT-2~\cite{brohan2023rt2visionlanguageactionmodelstransfer}, Octo~\cite{team2024octo}, and $\pi_{0.5}$~\cite{intelligence2025pi_} have shown remarkable success in robot control by directly learning mappings from observations and language instructions to robot actions. 
Pre-trained on large-scale internet data, these models generalize across tasks and embodiments, but operate as monolithic policies, making it difficult to adapt to new behaviors without requiring large amounts of additional data collection and fine-tuning. 
A common challenge in these models is to ground the language and vision components into the dynamics of the physical world.
To address this issue, reasoning can be separated from the control needed to actuate the real robot by leveraging skill libraries as an intermediate layer~\cite{zha2024distilling, wang2023voyager}. 
With this paradigm, the foundation model selects and parameterizes a skill from a set of predefined options. 
For example, SayCan~\cite{saycan2022arxiv} grounds language model plans in a robot's affordances by scoring available skills, while ProgPrompt~\cite{10161317} generates executable programs that compose skill primitives. 
Although effective for a known domain, the reliance on a finite set of skills limits generalization. 
Code as Policies~\cite{liang2022code} generates open-ended code at test time --- not relying on a set of pre-defined primitives --- while LMP~\cite{dai2026languagemovementprimitivesgrounding} argues that sufficiently general motion controllers, such as Dynamic Motion Primitives, enable general-purpose task execution when parameterized by VLMs.
However, even with zero-shot code and controller generation, the resulting behaviors may still fail to complete the desired task. 
Our skillbook addresses this gap by anchoring code generation in physical, user-provided corrections, while providing a continuously growing set of capabilities that extend beyond both the robot's original repertoire and what a foundation model can produce without grounded feedback.

\subsection{Skill Learning from Feedback and Experience}

Recent advances in embodied AI have explored how foundation models can be leveraged to construct, expand, and reuse skill libraries for long-horizon control. 
Agentic systems such as Voyager~\cite{wang2023voyager} iteratively generate executable programs, store successful behaviors in an ever-growing code-based skill library, and reuse them to solve novel tasks. 
Building on this paradigm \cite{tziafas2024lifelong} and \cite{zhao2024agentic} introduce mechanisms for composable skills via LLM-guided abstractions. 
Similarly, low-level behaviors can be gathered with reinforcement learning~\cite{dalal2025local} or derived from demonstrations~\cite{myers2024policy}, allowing high-level task and motion planners to sequence these new skills for zero- or few-shot transfer. 

Approaches such as DROC~\cite{zha2024distilling} distill language corrections from human users into a text-based knowledge base of rules, constraints, and preferences that are retrieved for future tasks, while other methods use feedback to shape reward functions~\cite{yu2023language}.
However, because these approaches utilize local feedback merely to adjust existing skill parameters or task plans, they remain limited to the original set of skills.
In contrast, MEMO aggregates corrections across multiple users and different tasks in order to synthesize generalizable, parameterized code functions that can cover novel skills.

\subsection{Retrieval-Augmented Generation for Robotics}

Retrieval-augmented generation (RAG) has emerged as an efficient paradigm for grounding language model reasoning in external knowledge that would not otherwise fit into the context windows of the underlying LLMs. 
While largely applied in language models, recent work has adapted RAG to robotics for grounding an agent's decision-making in past experiences~\cite{zhu2024retrieval}.
PragmaBot~\cite{qu2026pragmatistrobotlearningplan} employs a VLM to self-reflect on action outcomes, storing summarized experiences in a long-term memory that can be retrieved for future tasks. 
DROC~\cite{zha2024distilling} similarly retrieves past corrections based on textual and visual task similarity to inform skill creation in novel settings. 
These methods retrieve natural language experiences, corrections, or summaries to augment and contextualize the execution of desired tasks.
ELLMER~\cite{MonWilliams2025}, on the other hand, retrieves code examples from a curated knowledge base, demonstrating the effectiveness of retrieving executable code rather than just natural language.
However, unlike methods that rely on purely text-based memories or pre-curated static codebases, MEMO actively evolves its knowledge base. 
By clustering raw human feedback, our skillbook dynamically synthesizes executable, parameterized code templates that can be retrieved to guide skill generation.

\section{Problem Statement} \label{sec:problem}

We explore settings where a robot arm is performing tabletop manipulation tasks.
These include simple pick-and-place tasks (e.g., ``put the banana on the plate'') as well as more complex interactive tasks (e.g., ``toast food'').
A human gives the robot their natural language description of the task $\tau$, and the robot arm must convert this description into successful task execution.

\p{Scene Graph}
Let $o \in \mathcal{O}$ be the robot's observation of its environment state.
In our setting $o$ contains the robot's end-effector pose and an RGB-D image of the tabletop.
From this observation, the robot autonomously extracts a $3$D scene graph $\mathcal{G} = (\mathcal{V}, \mathcal{E}, o)$ which includes both the raw observation and its processed nodes and edges.
Each node $v \in \mathcal{V}$ contains an object's semantic label (e.g., ``door handle'') and that object's pose within the robot's coordinate frame.
Edges $e \in \mathcal{E}$ describe any natural language relationships between objects (e.g., ``the toaster door is closed'').
In our real-world experiments the robot builds this scene graph $\mathcal{G}$ using Gemini-Robotics ER \cite{team2025gemini} and LangSAM \cite{liu2024grounding, kirillov2023segment}.

\p{Policy}
The robot's policy converts the user's task description $\tau$ and the scene graph $\mathcal{G}$ into action $a \in \mathcal{A}$:
\begin{equation} \label{eq:P1}
    a \sim \pi(\cdot \mid \tau, \mathcal{G}, \rho)
\end{equation}
Here $\rho$ is a prior that guides the policy.
If the policy is a foundational robotics model this prior could be a set of pretrained weights \cite{intelligence2025pi_, kim2026cosmos, kim2024openvla}.
Alternatively, for neuro-symbolic policies the prior is a general-purpose prompt that explains how $\pi$ should function \cite{kwon2024language, kwon2025llmtamp}.
In our approach we instantiate the policy as a vision-language model, and $\rho$ is a task-invariant system prompt that initializes the policy.

\p{Skills}
The actions output by the policy are not necessarily position or torque commands.
Instead of these low-level behaviors, related works output trajectory snippets \cite{intelligence2025pi_}, motion primitives \cite{dai2026languagemovementprimitivesgrounding}, or function calls \cite{kwon2024language}.
In this paper we treat \textit{skills} as actions.
We define a skill as a parameterized function that intuitively grounds language reasoning into low-level motion control.
For example, \texttt{grasp(pose)} could be a skill that moves the robot's end-effector to \texttt{pose} and then closes its grippers.
Under this framework each action $a = (k, \theta)$ separates into a skill $k$ and the parameters $\theta$.
The set of all skills $k \in \mathcal{K}$ is the skill library, and the action space $\mathcal{A}= \mathcal{K} \times \Theta$ incorporates this skill library and its parameters $\theta \in \Theta$.
We emphasize that the robot's policy in \eq{P1} is constrained to the library: if the policy does not have access to a skill or sequence of skills that are necessary to complete the current task, the robot will inevitably fail.

\p{Feedback}
Ideally the robot's policy $\pi$ is able to complete the user's task $\tau$ on its first attempt by choosing the correct skills and parameters for those skills.
But in practice the robot will sometimes fall short.
When these failures occur, the user can provide text descriptions that explain what went wrong (e.g., ``no, you need to grasp the handle from the side'').
The robot should leverage this feedback to improve its future behavior.
Naively, we can just incorporate the human's text into the prior $\rho$, thereby helping the robot address a specific error.
But our goal is for the robot to make long-term improvements to $\pi$.
Instead of simply recalling the human's text for one task, we want the robot to build a skill that can be leveraged across multiple tasks --- e.g., an \texttt{open\_door()} function. 
Based on the human's short-term and localized corrections, our goal is to extract generalized feedback that the robot can leverage to dynamically expand $\mathcal{A}$ and increase its capabilities.

\begin{figure*}[t]
    \centering
    \includegraphics[width=\linewidth]{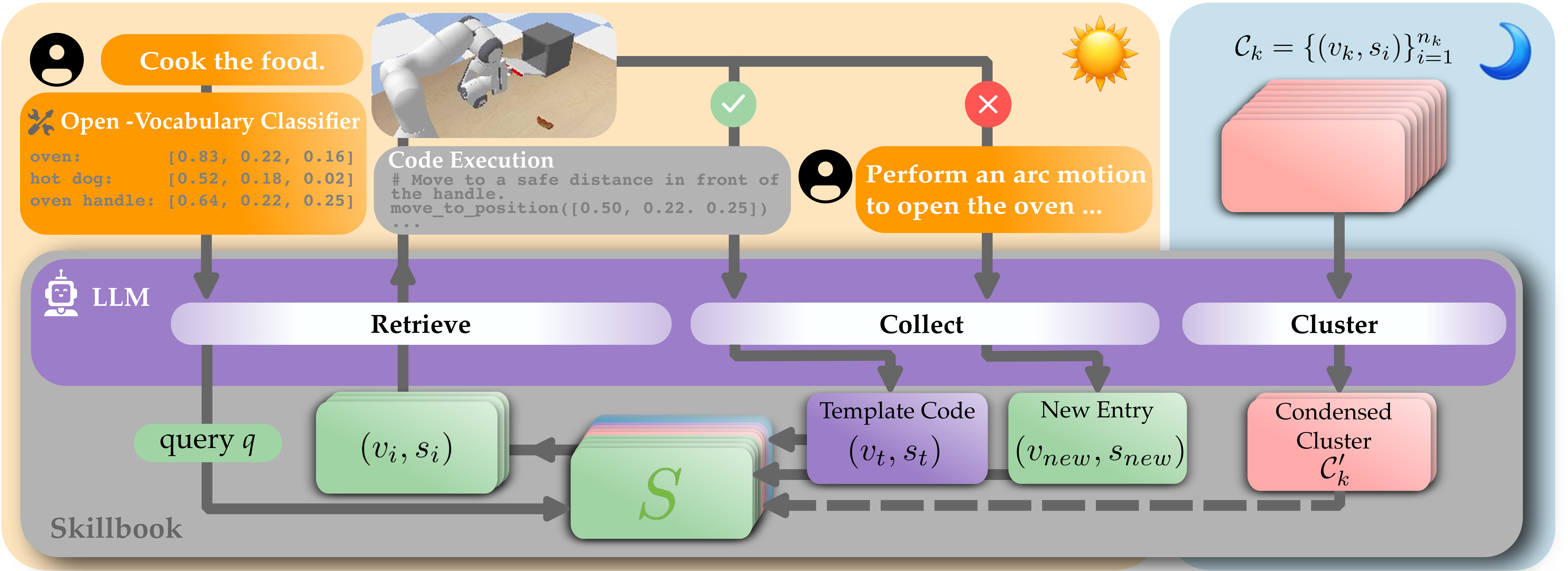}
    \vspace{-1.5em}    
    \caption{
    MEMO builds generalizable skills by retrieving, collecting, and clustering human feedback. 
    MEMO first decomposes a task into its atomic actions (i.e., subtasks). 
    Before selecting an action, it queries the skillbook for any subtask-relevant feedback or functions that it can reference.
    If the user intervenes, MEMO stores a paraphrased form of their feedback in the skillbook for future reference. 
    Otherwise, we store the action as a skill template.
    Offline MEMO clusters and compresses the entries in the skillbook to mitigate information loss and gather more general skills for future action generation. 
    }
    \label{fig:method}
    \vspace{-1.5em}
\end{figure*}

\section{MEMO: Memory Enhanced Manipulation} \label{sec:method}

To convert local feedback into general skills we propose \textbf{MEMO}.
Let $\mathcal{S}$ be a knowledge base of natural language corrections, general feedback, and skill templates.
We will refer to $\mathcal{S}$ as a \textit{skillbook}.
In its most basic form, MEMO is a method to build and maintain the skillbook $\mathcal{S}$ so that the robot's neuro-symbolic policy can expand its action space across multiple tasks.
We re-define \eq{P1} to condition $\pi$ on both the fixed prior $\rho$ and the adaptive skillbook $\mathcal{S}$:
\begin{equation} \label{eq:M1}
    a \sim \pi(\cdot \mid \tau, \mathcal{G}, \rho, \mathcal{S})
\end{equation}
At run time, the robot's policy retrieves relevant language feedback or function templates from $\mathcal{S}$, and then reasons across this information when generating code for parameterized skills $a$.
In Section~\ref{sec:M1} we explain the contents of $\mathcal{S}$ and how the robot uses human feedback and task success to fill its skillbook.
Next, in Section~\ref{sec:M2} we describe how the policy $\pi$ in \eq{M1} retrieves entries from $\mathcal{S}$ that are relevant for the current task.
Finally, in Section~\ref{sec:M3} we outline how the skillbook is clustered to extract more generalized entries from multiple pieces of local feedback.
Together, this process of collecting, retrieving, and clustering corrections into a skillbook forms our MEMO framework.
We visualize this MEMO approach in \fig{method}.

\subsection{Collecting Feedback into Skillbook Entries} \label{sec:M1}

We instantiate the knowledge gathered from human feedback as a vector database.
Here $\mathcal{S} = \{\left(v, s\right)\}$, where each $v$ is an embedding vector and $s$ is the associated skill information.
Intuitively, we can think of $v$ as the key used to identify the entry, and $s$ as the information stored within that entry.

\p{Parsing Human Feedback}
During task execution a human can stop the robot at any time and provide natural language feedback about its mistakes.
Of course, different users express the same ideas with different words, phrases, and syntax.
The robot therefore paraphrases the human's feedback with its language model to remove any information that is too task-specific, redundant, or irrelevant.
For instance: if the human says ``the robot should place its end-effector at position $(0.5, -0.3, 0.2)$ to grab the handle,'' the paraphrased text could be ``move to the door handle.''

Each piece of human feedback becomes a new entry in $\mathcal{S}$.
Specifically, the paraphrased text is the information $s$, and the embedding vector $v$ is set based on the robot's context.
For our tabletop manipulation tasks we divide this context into two parts: the natural language \textit{action} that the robot is trying to perform, and the \textit{object(s)} that the robot is interacting with.
If the human provides feedback when the robot is, for example, trying to open the toaster, the action token could be ``open'' and the object token could include ``toaster'' and/or ``door.''
Together these action and object tokens form the embedding vector $v = \left(v_{act}, v_{obj}\right)$.
%Optionally, designers can append the current scene graph $\mathcal{G}$ to $v$ to aid in state-conditioned retrieval.

When paraphrasing the human's task-specific feedback we also prompt the language model to try and extract higher-level corrections.
For example, if the user interrupts with statements like ``you hit the table while you were moving,'' then the language model may identify more high-level guidance such as ``ensure the robot stays a safe height above the table.''
The advantage of high-level feedback is that it can be applied across multiple contexts, and is not necessarily tied to the current action or objects.
We capture this task invariance within the skillbook $\mathcal{S}$ by indexing general statements $s$ with a shared global key $v_g$.

\p{Leveraging Task Success as Implicit Feedback}
User feedback tells the robot about its \textit{incorrect} motions.
But we can also reinforce \textit{correct} behaviors by storing successful code within the skillbook $\mathcal{S}$.
When the robot successfully completes a subtask, the robot converts the code it wrote for that subtask into a function template.
Here \textit{code} refers to the skills $k \in \mathcal{K}$ the robot invoked, the order of those skills, and any chosen skill parameters $\theta \in \Theta$.
The robot then takes the code for $(a_1, a_2, \ldots)$ and queries its language model for a more general form; i.e., removing any hardcoded values.
The inputs to this function template can be the environment scene, the current robot configuration, and the positions of task-relevant objects.
The template is then stored as a new entry $\left(v_f, s_f\right)$.
The language-generated code template is $s_f$, and the vector for retrieving that code is $v_f$.
As before, $v_f$ includes the action the robot used that code to complete, as well as any objects the code template interacts with.

\subsection{Retrieving Relevant Skillbook Entries} \label{sec:M2}

In Section~\ref{sec:M1} we explained how our skillbook is populated from user feedback and task success.
The size of this skillbook $\mathcal{S}$ grows over time as the robot attempts new manipulation tasks; as such, we cannot expect the robot to reason across its entire database when choosing actions in \eq{M1}.
Recent studies have shown that language models experience performance degradation as their input length increases \cite{du2025context, zhou2025gsm}. 
We therefore apply a retrieval-augmented generation (RAG) approach to collect the most relevant entries in $\mathcal{S}$ based on the robot's task $\tau$ and context $\mathcal{G}$.
These relevant entries are then entered into the current conversation.
Importantly, \textit{the robot does not simply copy any skill templates retrieved from $\mathcal{S}$} --- instead, we use these templates as a reference, and the robot constructs new code to generate embodied motions based on the system prior $\rho$ and retrieved entries in skillbook $\mathcal{S}$.

\p{Retrieving} 
Within our neuro-symbolic policy the robot first uses a vision and language model to reason across the task $\tau$.
The policy next determines its semantic sequence of subtasks: e.g., when asked to ``toast food'' the first subtask might be to ``open the toaster.''
At this point the policy in \eq{M1} performs retrieval to search for relevant entries in its knowledge base $\mathcal{S}$.
More formally, the policy compares the action and object(s) it plans to interact with against the embeddings $v$ in the skillbook:
\begin{equation*}
    r(q, v) = \frac{1}{\lambda_1 + \lambda_2} \Big( \lambda_1 \cos{(q_{act}, v_{act})} + \lambda_2 \cos{(q_{obj}, v_{obj})} \Big)
\end{equation*}
Here $q=\{q_{act}, q_{obj}\}$ is the robot's query (i.e., the actions and objects in its intended subtask) and $\cos$ is the cosine similarity.
We weight this similarity by hyperparameters $\lambda_1,  \lambda_2 > 0$.
Increasing the value of $\lambda_1$ means that the robot is more likely to retrieve entries with synonymous actions: e.g., if the subtask calls for ``grasping,'' the system might retrieve entries in $\mathcal{S}$ that ``grab,'' ``pick up,'' or ``hold.''
The retrieved information is then added to the context window for policy $\pi$.
This matches our framework in \eq{M1}: the robot selects actions while reasoning across its static system prompt $\rho$ and the relevant skillbook entries $(v, s) \in \mathcal{S}$.
The robot then generates code to complete its current subtask; this code is based on the templates in $\mathcal{S}$ and the skills in $\rho$.

\begin{table}[t]
    \centering
    \footnotesize
    \renewcommand{\arraystretch}{1.0}
    \caption{Experimental tasks, task types, and the total amount of human feedback collected for MEMO and DROC$-$V.}
    \vspace{-1.0em}
    \label{tab:us-tasks}
    \begin{tabular}{rcc}
        \hline
        \textbf{Task Name} & \textbf{Type} & \textbf{\# Feedback} \\
        \hline
        
        \rowcolor{gray!20} \multicolumn{3}{l}{\textbf{Test Tasks (used to collect user feedback)}} \\
        \hline
        Pick the left can & SR & 2 \\
        \rowcolor{gray!10} Put the banana on the plate & SR & 2 \\
        Put the cube away & LH & 11 \\
        \rowcolor{gray!10} Put the food in the microwave & TR & 14 \\
        Put the food in the pan and & & \\
        place it in the microwave & SR & 22 \\
        \rowcolor{gray!10} Put the trash away & SR & 29 \\
        Stack the cubes & LH & 3 \\
        \rowcolor{gray!10} Turn on the faucet & TR & 2 \\
        Set the table & LH & 11 \\
        \rowcolor{gray!10} Clean up the table & LH & 23 \\
        Heat the food & SR & 16 \\
        \rowcolor{gray!10} Cook the food & SR & 2 \\
        Empty the fridge & LH & 23 \\
        \rowcolor{gray!10} Make toast & SR & 7 \\
        Move the lonely object to the others & SR & 5 \\
        \rowcolor{gray!10} Move the right can to the left can & SR & 2 \\
        Open the fridge & TR & 2 \\
        \rowcolor{gray!10} Open the bottle & TR & 4 \\
        Wipe the plate & CR & 7 \\
        \rowcolor{gray!10} Season the food & SR & 37 \\
        \hline
        
        \rowcolor{gray!20} \multicolumn{3}{l}{\textbf{Held-Out Tasks (used for evaluation)}} \\
        \hline
        Place the apple on the table & LH & -- \\
        \rowcolor{gray!10} Pour the can & TR & -- \\
        Close the bottle & TR & -- \\
        \rowcolor{gray!10} Empty the cabinet & SR & -- \\
        Put the food in the oven & CR & -- \\
        \hline
        
        \rowcolor{gray!10} \textbf{Total Feedback} & & \textbf{224} \\
        \hline
    \end{tabular}
    \vspace{-2em}
\end{table}

\subsection{Clustering Entries into Generalized Skills} \label{sec:M3}

From Section~\ref{sec:M2} the robot now has a way to retrieve relevant information and apply it to the current task.
But the point of our skillbook is not just to leverage \textit{localized} feedback for specific situations (e.g., using text about how to open the toaster the next time we encounter the same toaster).
Rather, we want to extract \textit{generalized} code and language that the robot can use to generate skills for a variety of tasks.
In addition, we need the skillbook to scale with an increasing number of users and tasks.
For instance, if the skillbook contains $50$ entries about how it should open a door, these entries quickly become repetitive, inefficient, and even contradictory.
We therefore \textit{cluster} the skillbook entries offline to refine $\mathcal{S}$ and aggregate generalized knowledge.

\p{Clustering}
Our first step in this offline phase is to group the skillbook entries by their embeddings.
Specifically, we form $K$ different clusters $\mathcal{C}_k=\{(v_k, s_i)\}^{n_k}_{i=1}$ by grouping all feedback entries $s_i$ that share the same embedding vector $v_k$.
Our goal is to condense this cluster into more compact and generalized guidance.
Returning to our example --- instead of $50$ descriptions of how to open $50$ different doors, we would rather have a small number of entries that cover all of these instances.
MEMO accomplishes this compression by passing each cluster $\mathcal{C}_k$ through a language model to obtain a condensed feedback description $\mathcal{C}_k'$.
To address contradictory feedback and mitigate potential information loss, we condition the process on any related function templates $(v_f, s_f)$.
This design biases the clustering so that it prunes any feedback which does not agree with the function template --- and since the function template is built from successful code, we know that the human's guidance should be consistent with $s_f$.
The language model therefore inputs $\mathcal{C}_k$ and $(v_f, s_f)$, and outputs a new set of skillbook entries $\mathcal{C}_k'$ that combines similar entries while removing redundant or contradictory feedback.
Each entry in $\mathcal{C}_k'$ should align with the corresponding skill template and provide additional guidance not currently included within that template.
The size of $\mathcal{C}_k'$ is equal to or smaller than the size of the original $\mathcal{C}_k$, both in terms of the number of entries and the number of characters within those entries.

\begin{figure*}[t]
    \centering
    \includegraphics[width=1.0\linewidth]{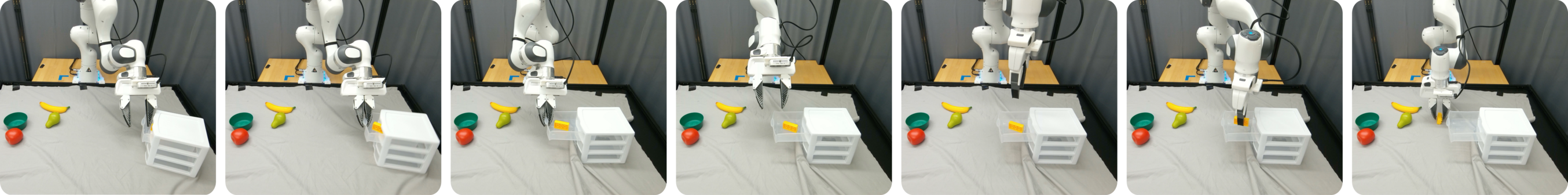}
    \caption{Our setup for real-world experiments. Here the robot uses MEMO to ``Empty the Cabinet'' in a zero-shot manner.}
    \label{fig:img_sequence}
    \vspace{-1.0em}
\end{figure*}

Clustering helps the skillbook generalize because it reasons across multiple pieces of related feedback and then provides a summarized version.
This also improves retrieval: instead of collecting $50$ ``opening'' entries with varying levels of usefulness, now the robot can retrieve a more compact set of mutually exclusive and aligned guidance for \eq{M1}.
Finally, because we condition clustering on templates from successful subtasks, the robot is able to prune erroneous feedback that may cause conflicts within the original skillbook.

\begin{figure}
    \centering
    \includegraphics[width=1.0\linewidth]{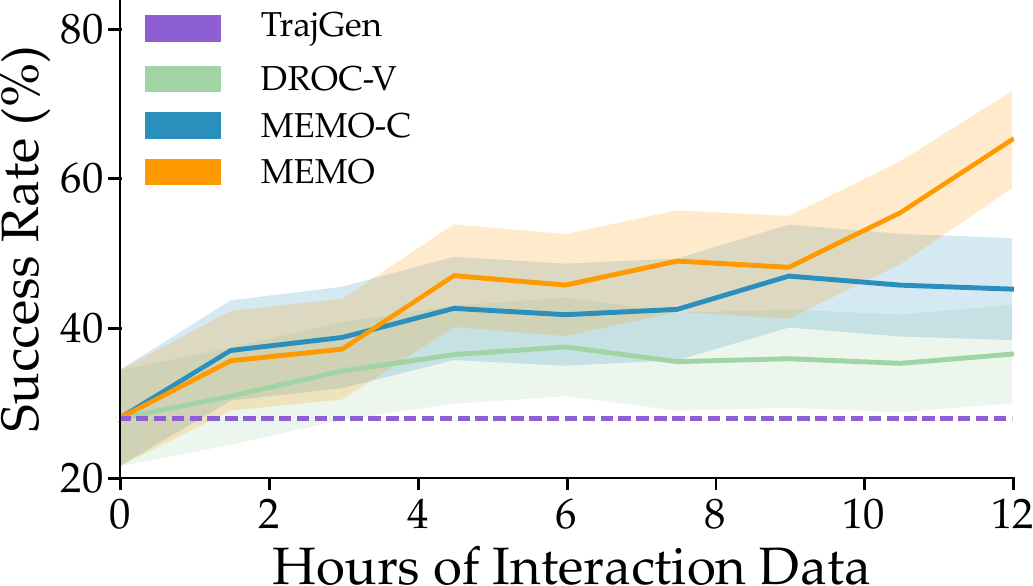}
    \caption{Zero-shot success rate for our held-out evaluation tasks in simulation. The $x$-axis shows the size of the skillbook in terms of user hours. Note that as the skillbook grows, the performance of MEMO$-$C and DROC$-$V tends to stagnate without clustering local corrections into generalized guidance. The average zero-shot success rate at the onset of the study (with no entries in the skillbook) is approximately 30$\%$, while as the skillbook grows in size, MEMO approaches a success rate of 80$\%$.}
    \label{fig:sims1}
    \vspace{-2.5em}
\end{figure}

\section{Experiments} \label{sec:experiments}

In this section we aim to answer three main questions in order to assess the capabilities of MEMO: 
1) We examine whether incorporating entries retrieved from the skillbook improves zero-shot generalization to unseen task configurations;
2) We evaluate whether our clustering approach helps MEMO better understand tasks and resolve conflicting feedback while maintaining a compact skillbook;
3) We investigate cross-task learning by examining how new skills are formed in the skillbook in a real-world evaluation.

\p{Evaluation Tasks}
We evaluate MEMO in a tabletop manipulation setting across simulated and real-world experiments using a 7-DoF Franka Emika Panda robot with a UMI gripper. 
In simulation we leverage $25$ tasks, twenty for creating our skillbook and the remaining five for evaluation. 
All $25$ tasks are listed in Table~\ref{tab:us-tasks}.
We specifically selected tasks designed to challenge the system across multiple aspects, including long-horizon planning (\textbf{LH}), contact-rich manipulation (\textbf{CR}), semantic reasoning about objects and task goals (\textbf{SR}), and coordinating both translational and rotational control (\textbf{TR}). 
We ensure that at least one task of each category is represented in the five evaluation tasks.  
\begin{figure}
    \centering
    \includegraphics[width=0.9\linewidth]{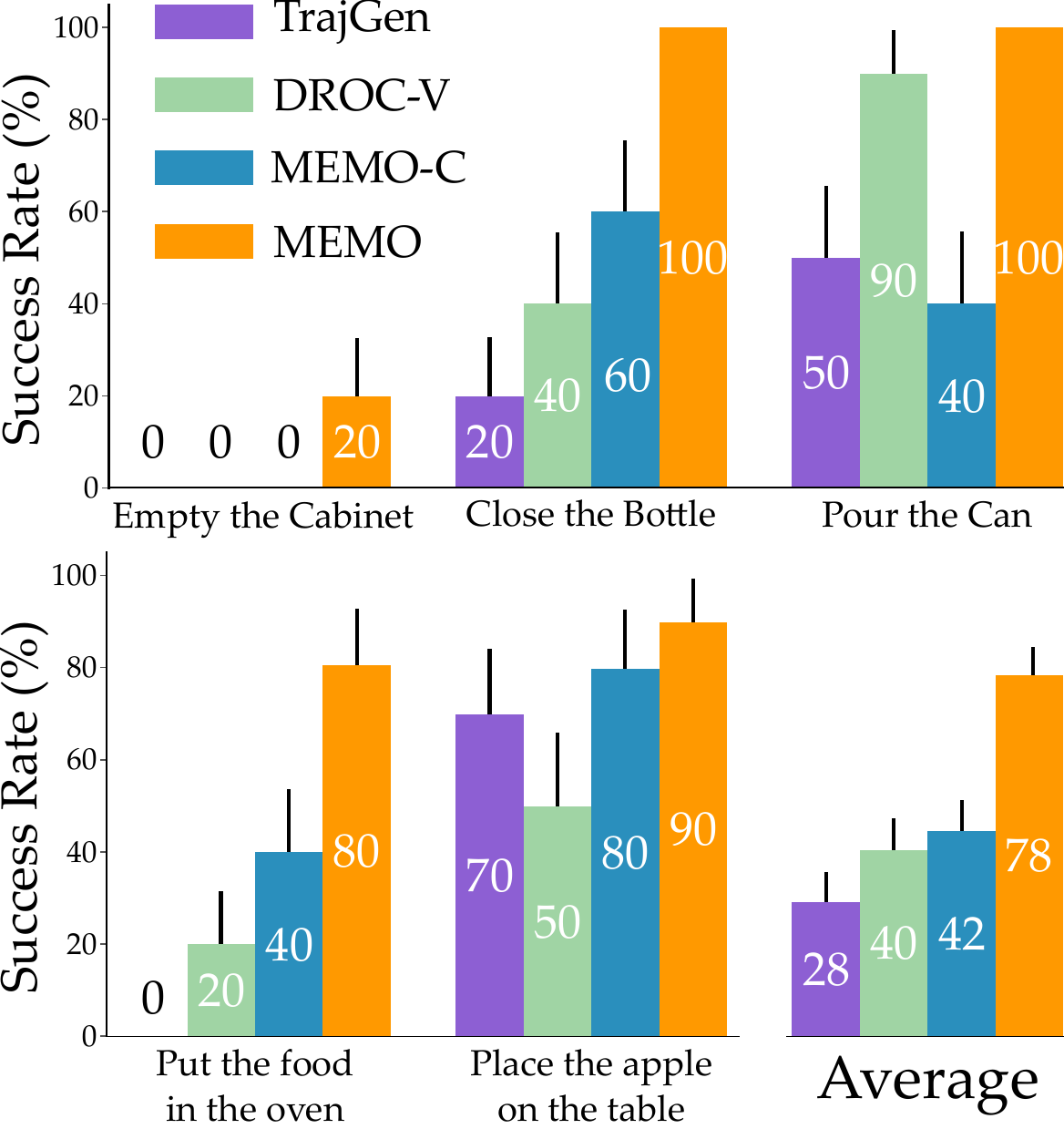}
    \caption{Breaking down \fig{sims1} with the complete $12$ hours of user feedback. Across the simulated evaluation tasks, MEMO achieves a higher zero-shot success rate than alternatives. Without the inclusion of offline clustering, MEMO is unable to generate the necessary skills for ``Empty the Cabinet'', ``Close the Bottle'', and ``Pour the Can.'' In unseen tasks MEMO achieves a success rate of $78\%$ as compared to $40\%$ for DROC$-$V.}
    \label{fig:all_sims}
    \vspace{-1.5em}
\end{figure}

We leverage the same five evaluation tasks in the simulated experiments and in the real world. 
The robot's control interface is identical between the simulated and real-world setups, i.e., the functions that control the robot require the same input parameters. 
For perception in the real world, we utilize an Orbbec Femto Mega RGB-D camera as well as two Intel RealSense D435 cameras for training and deployment of $\pi_{0.5}$, one mounted on the robot's wrist and one positioned as a side view camera.
The side camera is configured to approximately match the field of view of the main Orbbec camera.
Given these camera feeds, we leverage Gemini Robotics-ER 1.5~\cite{team2025gemini} to detect objects and LangSAM~\cite{kirillov2023segment, liu2024grounding} to obtain fine-grained segmentation masks to reconstruct 3D object bounding boxes. 
Finally, for specific object components (e.g., a cabinet handle) we attach an ArUco marker to estimate pose in $\text{SE}(3)$ space.
We use \textit{gemini-3-flash-preview} as our large language model and \textit{all-MiniLM-L6-v2} as our sentence transformer. %for neuro-symbolic approaches.

\begingroup
\renewcommand{\arraystretch}{1.0}
\begin{table*}
    \centering
    \small
    \setlength{\tabcolsep}{0.5em}%
    \caption{Performance on real-world tasks. We report the success percentage and the average amount of feedback per task $(\#)$. We find that \textbf{MEMO} reaches the highest overall performance with the least amount of feedback during evaluation.
    }
    \label{tab:results}
    \begin{tabular}{ r|cccc}
    \toprule
    \multicolumn{1}{r|}{\raisebox{0.3em}{\textbf{Task Name}}}
    & \raisebox{0.3em}{\textbf{MEMO}}
    & \raisebox{0.3em}{\textbf{MEMO$-$C}}
    & \raisebox{0.3em}{\textbf{DROC$-$V} \cite{zha2024distilling}}
    & \raisebox{0.3em}{$\pi_{0.5}$ \cite{intelligence2025pi_}}
    \\
    \midrule
    Place the apple on the table
    & $\mathbf{100}~(0.2)$ & $\mathbf{100}~(0.6)$ & $80~(1.4)$ & $20$
    \\
    Pour the can
   & $\mathbf{100}~(0.2)$ & $\mathbf{100}~(1.2)$ & $60~(1.8)$ & $20$ \\
    Close the bottle
    & $\mathbf{100}~(1.0)$ & $\mathbf{100}~(2.0)$ & $80~(2.2)$ & $0$ \\
    Empty the cabinet
    & $\mathbf{80}~(2.8)$ & $\mathbf{80}~(3.2)$ &$60~(3.8)$ & $0$\\
    Put the food in the oven
    & $\mathbf{60}~(3.4)$ & ${40}~(4.0)$ & $20~(4.6)$ & $20$ \\ %\midrule
\midrule
\multicolumn{1}{r|}{\textbf{Overall Performance}} 
& $\mathbf{88}~\mathbf{(1.52)}$ & 84~(2.2) & 60~(2.76) & 12\\
\bottomrule
     \end{tabular}
\vspace{-1.5em}
\end{table*}
\endgroup

\p{Metrics}
To assess the performance of all methods across simulation and real-world tasks, we report the overall task success rate across five trials. 

\p{Baselines and Ablations}
We compare MEMO against state-of-the-art baselines, as well as ablated versions of MEMO. 

Our primary baselines are DROC$-$V \cite{zha2024distilling} and $\pi_{0.5}$~\cite{intelligence2025pi_}.
DROC$-$V is a neuro-symbolic approach that stores human feedback to apply context-specific corrections.
Intuitively, we can think of DROC$-$V as similar to MEMO, but without the code templates, general feedback, and clustering within our method.
We provide the same set of human feedback to DROC$-$V as we use to build our skillbook. 
The baseline $\pi_{0.5}$ is a representative vision-language-action (VLA) model. 
We train $\pi_{0.5}$ on a set of real-world demonstrations composed of all $25$ tasks utilized in simulation for $20000$ steps, following the officially released Libero fine-tuning scheme~\cite{intelligence2025pi_}.

We also ablate our method to test the following versions:
\begin{enumerate}[leftmargin=*]
    \item {\textbf{MEMO$-$C:}}
    We test our method without the \textit{clustering} stage. This enables us to explore whether clustering helps MEMO obtain more generalized skills.

    \item {\textbf{MEMO$-$S} (TrajGen):}
    We test our method without the use of any skillbook. Effectively, this version of MEMO is equivalent to \textbf{TrajGen}~\cite{kwon2024language}, but with the inclusion of an open vocabulary classifier. 
\end{enumerate}

\subsection{Building the Skillbook: Human Subject Study} \label{sec:E1}

MEMO is centered around the creation of a skillbook that accumulates and clusters feedback from users in order to improve its capability to generalize to novel tasks. 
To this end, we utilize our $20$ training tasks in simulation and ask $20$ human subjects to provide free-form natural language feedback to improve the robot's behavior.
The $20$ participants are made up of $7$ females and $13$ males (age $24 \pm 5.5$) recruited from the Virginia Tech community; some participants chose to participate in the study multiple times.
Eleven participants report familiarity with the use of large-language models ``multiple times per week.''
Seven of the $20$ participants report having no robotics experience prior to the study. 
All users provide informed written consent consistent with IRB \#23-1237 and were compensated for their time.

Participants observe the robot's motion as part of the current subtask to a) determine if the robot succeeded in its task or b) provide unconstrained natural language feedback that they believe will improve task performance. 
The users could interrupt the robot at any time during task execution and provide feedback.
When they did, their feedback was paraphrased by the language model, labeled with an \textit{action}-\textit{object(s)}-\textit{scene} triplet corresponding to the current subtask, and added to the skillbook. 
The environment was then reset to the beginning of the respective subtask and the robot attempted to complete the task again, now with the previously provided feedback as a part of its skillbook.
Table~\ref{tab:us-tasks} lists the total number of feedback entries provided for each task.
Each session lasted $30$ minutes.
During these $30$ minutes, participants worked through as many tasks as possible following the order shown in Table~\ref{tab:us-tasks}; each new participant picked up from where the previous participant left off. 
With this protocol the $20$ participants provided three rounds of feedback.
We found that $13$ participants were required for the first iteration of all $20$ tasks, $4$ participants were able to complete the second iteration, and only $3$ participants were needed for the third and final iteration.

\subsection{Zero-Shot Generalization}
\label{sec:exp-sim}

Now that we have the skillbook collected in Section~\ref{sec:E1}, we are ready to test our first hypothesis --- whether retrieved entries from the skillbook improve zero-shot performance.
We evaluate in simulation on five held-out evaluation tasks (see Table~\ref{tab:us-tasks}) for which MEMO has no prior feedback.
We compare MEMO against three baselines: {MEMO$-$C}, {MEMO$-$S} ({TrajGen}), and {DROC$-$V}.
During this evaluation no methods received feedback on the held-out tasks. 
\fig{sims1} shows that the addition of our skillbook consistently improves performance ($78\%$ task success) over TrajGen and DROC$-$V ($40\%$ and $28\%$, respectively).
DROC$-$V has been trained by providing the same set of feedback for its offline storage (without our templated code generation), while TrajGen does not utilize any feedback (thus resulting in a horizontal line).
We observe that our clustered skillbook increases performance --- particularly for large amounts of feedback past the $10$-hour mark --- demonstrating the importance of feedback aggregation ($36\%$ improvement).
Based on this zero-shot performance, we conclude that MEMO generalizes user feedback to similar tasks, and this generalization is due to the code templates and clustering.

\subsection{Skillbook Clustering} \label{sec:E3}

As seen in Section~\ref{sec:exp-sim}, clustering our skillbook significantly improves performance. 
To investigate \textit{why} clustering helps, we separately assess MEMO's success rate for all five held-out tasks (see \fig{all_sims}).
We observe that --- with the clustered skillbook --- MEMO is able to generate skills and complete tasks that it cannot regularly solve without this clustering (see $78\%$ average success rate over $42\%$ for the next-best method).
Consider the \textit{``Pour the Can''} task: when checking the skillbook, we find that clustering removes references that are not actually relevant to the task but may be erroneously retrieved (see $40\%$ vs. $100\%$ success rate).
Indeed, MEMO$-$C actually falls below several of the baselines in \fig{all_sims} because it retrieves incorrect or extraneous feedback from the skillbook, causing the policy to generate ineffective skills.
These results highlight that compression plays a critical role in MEMO beyond just its simplification of the relevant context; compression also resolves erroneous retrieval of unrelated and conflicting context.

\subsection{Real-World Evaluation} \label{sec:M4}

Thus far we have focused on simulation tasks for both constructing and testing the skillbook.
We finally take this same skillbook and apply MEMO to a real-world robot arm where we can explore the transferability of our approach and the creation of the relevant skills from the skillbook (see \fig{img_sequence}).
We test the same five held-out evaluation tasks that were used in simulation.
Table~\ref{tab:results} shows the task success rate of MEMO, including the MEMO$-$C, DROC$-$V, and $\pi_{0.5}$ baselines. 
Each task is attempted five times while allowing for up to five instances of feedback per attempt. 
We observe that MEMO provides similar or better performance to MEMO$-$C, while requiring significantly less feedback ($1.52$ pieces of feedback for MEMO as compared to $2.76$ for MEMO$-$C).
Similarly, MEMO significantly outperforms DROC$-$V in both performance and issued feedback, and is also more effective than $\pi_{0.5}$ in its overall performance. 
We emphasize that the feedback which MEMO uses was collected in simulation, and not the real world; hence, the fact that MEMO can extrapolate code for new and performant skills suggests that there is some cross-task and cross-environment skill transfer.

\section{Conclusion} \label{sec:conclusion}

Neuro-symbolic policies rely on skills to ground their semantic subtasks into robot motions.
To dynamically expand the skills that a robot arm can generate, we propose MEMO.
MEMO is based on aggregating human language corrections and successful code snippets into a retrieval-augmented skillbook.
Rather than simply storing and retrieving the most relevant entries, MEMO clusters the human's feedback while conditioning on successful code templates.
The robot then reasons across these generalized functions and text guidance at run time, enabling the policy to generate code for new skills while still being informed by human feedback.
Our experiments show that (a) using the skillbook improves generalization to unseen task configurations, (b) clustering the skillbook results in cross-task learning that forms new skills from existing ones, and (c) MEMO achieves a higher zero-shot success rate than neuro-symbolic approaches which only reason about relevant human feedback.

% \p{Limitations}

\balance
\bibliography{references} 
\bibliographystyle{ieeetr}

\end{document}